\documentclass[a4paper]{article}

\usepackage{INTERSPEECH2022}
\usepackage{amsmath}
\usepackage{multirow}
\usepackage{wrapfig}
\usepackage{url}

\title{Self-Normalized Importance Sampling for Neural Language Modeling}
\name{Zijian Yang$^1$, Yingbo Gao$^{1,2}$, Alexander Gerstenberger$^1$, \\Jintao Jiang$^2$, Ralf Schlüter$^{1,2}$, Hermann Ney$^{1,2}$}
\address{$^1$Human Language Technology and Pattern Recognition Group, Computer Science Department\\RWTH Aachen University, 52074 Aachen, Germany $^2$AppTek GmbH, 52062 Aachen, Germany}
\email{\{zyang|ygao|schlueter|ney\}@cs.rwth-aachen.de\\
alexander.gerstenberger@rwth-aachen.de, jjiang@apptek.com}

\begin{document}

\maketitle
\begin{abstract}
To mitigate the problem of having to traverse over the full vocabulary in the softmax normalization of a neural language model, sampling-based training criteria are proposed and investigated in the context of large vocabulary word-based neural language models.
These training criteria typically enjoy the benefit of faster training and testing, at a cost of slightly degraded performance in terms of perplexity and almost no visible drop in word error rate.
While noise contrastive estimation is one of the most popular choices, recently we show that other sampling-based criteria can also perform well, as long as an extra correction step is done, where the intended class posterior probability is recovered from the raw model outputs.
In this work, we propose self-normalized importance sampling.
Compared to our previous work, the criteria considered in this work are self-normalized and there is no need to further conduct a correction step.
Through self-normalized language model training as well as lattice rescoring experiments, we show that our proposed self-normalized importance sampling is competitive in both research-oriented and production-oriented automatic speech recognition tasks.
\end{abstract}


\section{Introduction}

Nowadays, word-based neural language models (LMs) consistently give better perplexities than count-based language models \cite{sundermeyer2012lstm,  irie19:trafolm}, and are commonly used for second-pass rescoring or first-pass decoding of automatic speech recognition (ASR) outputs \cite{6854535, beck2019lstm, beck2020lvcsr}.
One challenge to train such LMs, especially when the vocabulary size is large, is the traversal over the full vocabulary in the softmax normalization.
During both training and testing, this brings inefficiencies and calls for sampling-based methods to ease the calculation burdens.
Previously, many sampling-based training criteria are proposed and investigated.
Some prominent examples include: hierarchical softmax \cite{DBLP:journals/corr/abs-1301-3781}, negative sampling \cite{DBLP:conf/nips/MikolovSCCD13}, importance sampling (IS) \cite{pmlr-vR4-bengio03a}, and noise contrastive estimation (NCE) \cite{pmlr-v9-gutmann10a}.

While NCE is one of the most popular choices for language modeling \cite{DBLP:conf/icml/MnihT12, goldberger-melamud-2018-self, 7179005}, recently Gao et al. \cite{gao21b_interspeech} show that other sampling-based criteria can also perform well, as long as an extra correction step is done, where the intended class posterior probability is recovered from the raw model outputs.
The motivation of this work begins from there.
We notice that by introducing simple modifications to the original IS training criteria, the additional correction step can be omitted and the class posteriors can be directly obtained from the model outputs.
In other words, LMs trained with such criteria are also self-normalized - straightforwardly, we call this type of training: self-normalized importance sampling.

Compared to Gao et al. \cite{gao21b_interspeech}, our method is simpler in the sense that no additional correction is required, and the LMs are directly trained to give the intended class posteriors.
Compared to NCE, which is also a self-normalized criterion, our method is comparable and competitive, which we further show through extensive experiments on both research-oriented and production-oriented ASR tasks.

\section{Related Work}

Neural network LMs are shown to bring consistent improvements over count-based LMs \cite{sundermeyer2015:lstm, sundermeyer2012lstm, irie19:trafolm}.
These neural LMs are then used either in second-pass lattice rescoring or first pass decoding for ASR \cite{6854535, beck2019lstm, beck2020lvcsr, kumar2017lattice, li2021parallelizable}.
To mitigate the problem of having to traverse over the full vocabulary in the softmax normalization, various sampling-based training criteria are proposed and investigated \cite{DBLP:journals/corr/abs-1301-3781, DBLP:conf/nips/MikolovSCCD13, pmlr-vR4-bengio03a, pmlr-v9-gutmann10a, DBLP:conf/icml/MnihT12, goldberger-melamud-2018-self, 7179005}.
These methods mainly deal with how to build up a contrastive loss such that the model is able to tell true target words from noisy ones.
Related but originated from a different motivation, that is, to explicitly encourage normalization during training and hope for self-normalization during testing, variance regularization is also discussed in literature \cite{7533441}.
Recently, in \cite{gao21b_interspeech}, the authors identify the relationships between model optimums under various sampling-based training criteria and the contextual posterior probabilities.
They show that with one additional correction step, the intended posteriors can be recovered.
This work can be thought of as a follow-up and extention of \cite{gao21b_interspeech}, where we propose several simple modifications to the IS training criterion to directly enable self-normalized models, without having to do the additional correction step.

\section{Methodology}
In this section, we first describe several training criteria, including Binary Cross Entropy (BCE), NCE and IS.
Then, we pinpoint the reason why IS is not self-normalized by default.
Afterwards, we move on to discuss three modes of modifications to enable self-normalization of IS.
In the following discussions, we define $c$ as the running index in the target vocabulary size $C$, $x$ as the history for next word prediction and $n$ as the running index of word position $N$.
\label{sec:methodology}

\subsection{Binary Cross Entropy (BCE)}

BCE is a traditional training criterion, which only requires the model outputs $q_\theta(x,c)$ to be bounded in $(0,1)$, and not necessarily a normalized distribution.
\begin{equation}
    F_{\text{BCE}} = \frac{1}{N} \sum_{n=1}^N \{ \log q_\theta(x_n,c_n) + \sum_{c \neq c_n} \log ( 1 - q_\theta(x_n,c)) \}
    \label{eq: BCE}
\end{equation}
Given enough amount of training pairs $(x_n,c_n)$, it can be shown that the optimum $\hat{q}_\theta(x,c)$ when maximizing $F_{\text{BCE}}$ is the empirical posterior distribution $p(c|x)$, which indicates that the model outputs are self-normalized. 

\subsection{Fundamentals of Sampling-Based Training Criteria}
In sampling-based training criteria, for each training pair $(x_n,c_n)$, given the number of samples $K$ and the noisy distribution $D$ for sampling, the summation of some function $f(x_n,\Tilde{c}_k,\mathbb{E}(\mathbb{C}_{K}(c)))$ over $K$ random samples $\Tilde{c}_k$, can approximate the summation of $f$ over all the classes. The expected count $\mathbb{E}(\mathbb{C}_{K}(c))$ is the expectation of the number of the appearance of class $c$ in all samples. For instance, if we do sampling with replacement, the expected count would be $ KD(c)$. The term $\mathbb{E}(\mathbb{C}_{K}(c))$ in function $f$ is usually used to control the variance, i.e. with the increase of $K$, the variance goes down and the approximation is more precise. In the following discussions, to simplify the notation, we define 
\begin{equation}
    \mathbb{E}(\mathbb{C}_{K}(c)) \equiv E_c
\end{equation}
Omitting the summation over $n$ for simplicity, the approximation of the sampling method is:
\begin{equation}
\begin{aligned}
   \sum_{k=1}^K f(x_n,\Tilde{c}_k,&E_{\Tilde{c}_k}) \approx  \sum_{c=1}^C E_c \cdot f(x_n,c,E_c)
\end{aligned}
   \label{eq: sum_estimator}
\end{equation}
\subsection{Noise Contrastive Estimation (NCE)}

NCE is a popular BCE-style criterion, with the function $f$ in the following form:
\begin{equation}
    f(x_n,\Tilde{c}_k,E_{\Tilde{c}_k}) = \log( 1 - \frac{q_\theta(x_n,\Tilde{c}_k)}{q_\theta(x_n,\Tilde{c}_k) + E_{\Tilde{c}_k}})
\end{equation}
More specifically, the NCE criterion can be written as:
\begin{equation}
\begin{aligned}
    F_{\text{NCE}} &=  \frac{1}{N} \sum_{n=1}^N \{ \log \frac{q_\theta(x_n,c_n)}{q_\theta(x_n,c_n) + E_{c_n}} \\
    &\quad + \sum_{k=1}^K \log( 1 - \frac{q_\theta(x_n,\Tilde{c}_k)}{q_\theta(x_n,\Tilde{c}_k) + E_{\Tilde{c}_k}}) \}
\end{aligned}
\end{equation}
Here the first term in the summation over $n$ corresponds to the target class $c_n$ and the second term corresponds to the approximation shown in Equation (\ref{eq: sum_estimator}). With enough amount of training data and samples $K$, the optimum of model output is $\hat{q}_\theta(x,c) \approx p(c|x)$.

\subsection{Importance Sampling (IS)}
IS is another sampling-based BCE-style criterion.
\begin{equation}
    F_{\text{IS}} = \frac{1}{N} \sum_{n=1}^N \{\log q_\theta(x_n,c_n)+ \sum_{k=1}^K \frac{\log (1-q_\theta(x_n,\Tilde{c}_k))}{E_{\Tilde{c}_k}} \}
    \label{eq: IS}
\end{equation}
Here, with enough amount of training data and samples $K$, the model optimum is not normalized.
\begin{equation}
    \hat{q}_\theta(x,c) \approx \frac{p(c|x)}{1+p(c|x)}
\end{equation}
Nonetheless, as described in \cite{gao21b_interspeech}, $\hat{q}_\theta(x,c)$ can be transformed to $p(c|x)$ with an additional correction step. 

In IS, the function $f$ has a simpler form, which is the second log term in BCE criterion devided by the expected count.
\begin{equation}
     f(x_n,\Tilde{c}_k,E_{\Tilde{c}_k}) = \frac{\log (1-q_\theta(x_n,\Tilde{c}_k))}{E_{\Tilde{c}_k}}
\end{equation}
According to Equation (\ref{eq: sum_estimator}), the summation over samples is actually an approximation of the summation of the second log term in BCE over all classes,
\begin{equation}
    \sum_{k=1}^K f(x_n,\Tilde{c}_k,E_{\Tilde{c}_k}) \approx \sum_{c=1}^C \log(1-q_{\theta}(x_n,c))
    \label{eq: is_sum}
\end{equation}
which makes IS very similar to the original BCE criterion, except that the sum is over all classes rather than excluding the target class (notice the summation over $c$ in Equation (\ref{eq: is_sum}) and (\ref{eq: BCE})). Thus, our motivation is to modify the original IS criterion to have an approximation of the original BCE criterion, in order to directly obtain $p(c|x)$ as the optimum output, hence achieving self-normalized models.

\subsection{Self-Normalized Importance Sampling}
\subsubsection{Sampling including the Target Class}
\label{sec: Mode1}
When the target class is allowed to be sampled, from the discussion above, the model is not self-normalized. Therefore, an intuitive way to exclude the target class from the summation is to simply substract the target term $\log (1-q_\theta(x_n,c_n))$.
\begin{equation}
\begin{aligned}
    F_{\text{Mode1}} &= \frac{1}{N} \sum_{n=1}^N \{\log q_\theta(x_n,c_n)+ \sum_{k=1}^K \frac{\log (1-q_\theta(x_n,\Tilde{c}_k))}{E_{\Tilde{c}_k}} \\
    &\quad -\log(1-q(x_n,c_n))\} \\
    & \approx F_{\text{BCE}}
\end{aligned}
\end{equation}
With this simple modification, the sampling-based criterion is exactly an approximation of BCE criterion and the model optimum is now at $p(c|x)$. However, if the target class is not sampled, the additional $-\log(1-q(x_n,c_n))$ term can result in a large gradient. When target class $c_n$ does not appear in $\Tilde{c}_k$, we can consider the gradient with respect to $\log q(x_n,c_n)$ of a single sentence $n$.
\begin{equation}
    \frac{\partial F_{\text{Mode1}, n}}{\partial \log q(x_n,c_n)} = 1 + \frac{q(x_n,c_n)}{1 - q(x_n,c_n)}
\end{equation}
Suppose that $p(c_n|x_n)$ is a large number (close to one) and $q_\theta(x_n,c_n)$ is close to its optimum $p(c_n|x_n)$, the gradient is large and makes it hard to converge. In the following experimental results, we show that when $K$ is small, which means there is less chance to sample the target class out, this can lead to a bad performance. 

\subsubsection{Sampling excluding the Target Class}
Another way to obtain the self-normalized output is to find some proper distribution or function $f$ to directly approximate the summation without target class $c_n$. 
\begin{equation}
    \sum_k f(x_n,\Tilde{c}_k,E_{\Tilde{c}_k} ) \approx  \sum_{c \neq c_n} \log (1 - q_\theta(x_n,c))
\end{equation}

Considering equation  (\ref{eq: sum_estimator}), we propose two approaches: letting the expected count of $c_n$ be zero, i.e. $E_{c_n}=0$, or having a function $f_{c_n}$ that is specific for each target class and letting the value of the function be zero when the summation index $c$ is $c_n$, i.e. $ f_{c_n}(x_n,c_n, E_{c_n})=0$. 

For the first approach (Mode2), $E_{c_n}=0$ is equivalent to $D(c_n)=0$, i.e. the probability to sample the target class being zero. To this end, we utilize different distributions $D_{c_n}(c)$ with $D_{c_n}(c_n)=0$ for different target classes during training. Compared to IS, Mode2 has the same formula, but different noise distributions for sampling.
\begin{equation}
\begin{aligned}
    F_{\text{Mode2}} &= \frac{1}{N} \sum_{n=1}^N \{\log q(x_n,c_n)+ \sum_{k=1}^K \frac{\log (1-q(x_n,\Tilde{c}_k))}{E_{\Tilde{c}_k}} \}\\
\end{aligned}
\end{equation}
Because in this way it is not possible to sample the target class $c_n$, there is no devision-by-zero problem in the criterion.

For the second approach, we simply set $f_{c_n}$ to zero when the target class gets sampled out.
\begin{equation}
     F_{\text{Mode3}} = \frac{1}{N}\sum_{n=1}^N \{\log q(x_n,c_n)+ \sum_{k=1}^K f_{c_n}(x_n,\Tilde{c}_k,E_{\Tilde{c}_k})\}
\end{equation}
where
\begin{equation}
    f_{c_n}(x_n,\Tilde{c}_k,E_{\Tilde{c}_k}) = \left \{\begin{array}{rl}
        0 & \text{if} \quad \Tilde{c}_k = c_n \\
        \frac{\log (1-q(x,\Tilde{c}_k))}{E_{\Tilde{c}_k}} & \text{else}
    \end{array} \right .
\end{equation}
Compared to Mode2, Mode3 is more efficient since it can use one distribution for all training pairs $(x_n,c_n)$. For Mode2, on the other hand, it is necessary to use different distributions for different target classes. For Mode3, it is safer to do sampling without replacement, since if replacement is allowed, in extreme cases there can be many samples where $f$ evaluates to zero, which adds no contribution to the summation and thus can influence the precision of the approximation.

\section{Experimental Results}

\subsection{Experimental Setup}

We conduct experiments on public datasets including Switchboard and Librispeech. Additionally, we experiment with one in-house dataset: 8kHz US English (EN). The detailed statistics of these three datasets can be found in Table \ref{tab:data_stat}. For EN, 7473M in Table \ref{tab:data_stat} is the total amount of running words used for baseline count-based LM training, while roughly 694M running words are used for neural network training.

We follow the setup in \cite{gao21b_interspeech} in terms of model architectures. For Switchboard and EN, we use LSTM LMs with 2 layers, 2048 hidden dimension and 128 embedding dimension. For Librispeech, we use Transformer \cite{vaswani2017attention} LMs, with 42 layers, 2048 feed-forward hidden dimension, 512 key/query/value dimension, 8 attention heads, and 512 input embedding dimension, following the architecture in \cite{irie19:trafolm}. The acoustic models are hybrid hidden Markov models, with 7 bidirectional LSTM layers and 500 dimension. We refer the reader to \cite{kitza2019cumulative} and \cite{luscher2019rwth} for more details. 

We evaluate our models with perplexities (PPLs) and second-pass rescoring word error rates (WERs). For Switchboard and EN, the lattices for rescoring are generated with 4-gram Kneser-Ney LMs \cite{kitza2019cumulative}. For librispeech, the lattice is generated with a LSTM LM \cite{beck2019lstm}. All the model outputs are normalized properly when computing PPLs in order to have a fair comparison. In rescoring, the outputs of models trained with normalized or self-normalized criteria are used as is and no additional renormalization is done. For Importance Sampling, as the output cannot directly be used as scores, we obtain scores by a proper mapping of outputs \cite{gao21b_interspeech}. We report rescoring results on clean and other test sets for Librispeech and on Switchboard (SWB) and CallHome (CH) Hub5'00 test sets for Switchboard. For EN, we use an in-house test set. 


For sampling-based criteria, we use log-uniform distributions as noise distributions, following \cite{gao21b_interspeech}. If not mentioned otherwise, the number of samples $K$ is $8000$. We do sampling with replacement, except Mode3. During training, each batch shares the same sampled classes for efficiency \cite{zoph2016simple, jozefowicz2016exploring}, except Mode2. For Mode2, to build up the desired distribution for each training pair $(x_n,c_n)$, we use $C-1$ as the number of labels in log-uniform, and map the sampled indices $\Tilde{c}'_k$ to $\Tilde{c}_k$.
\begin{equation}
    \Tilde{c}_k = \left \{\begin{array}{rl}
        \Tilde{c}'_k & \text{if} \quad \Tilde{c}'_k < c_n \\
        \Tilde{c}'_k + 1  & \text{else}
    \end{array} \right .
\end{equation}

The average training time per batch is evaluated on NVIDIA GeForce GTX 1080 Ti GPUs, with a batch size of 32 for Switchboard, 64 for LibriSpeech and EN.

\begin{table}[]
\centering
\caption{Data statistics of the three datasets.}
\begin{tabular}{|l|r|r|}
\hline
\multicolumn{1}{|c|}{corpus} & \multicolumn{1}{c|}{vocabulary size} & \multicolumn{1}{c|}{running words} \\ \hline \hline
Switchboard   &   30k    &     24M        \\ \hline
Librispeech   & 200k       &      812M          \\ \hline
EN        &       250k    &   7473M   \\ \hline
\end{tabular}

\label{tab:data_stat}
\end{table}

\subsection{Main Results}

We first compare three modes of self-normalized IS on Switchboard, employing NCE as the baseline. As shown in Table \ref{tab: swb_compare}, for Mode1,  when $K=100$, the performance is much worse than other methods. When $K=8000$, on the other hand, the performance is on par with others, which verifies our statement about the big gradient problem in Section \ref{sec: Mode1}. For both $K=100$ and $K=8000$ in Mode2 and Mode3, the performance is comparable with NCE. Considering that Mode3 is more efficient than Mode2 in terms of the noise distribution, we choose Mode3 to be our main method. For both Mode2 and Mode3, we contribute the relatively small difference in PPLs to the small vocabulary size.

\begin{table}[h!]
\center
\caption{PPLs of NCE and Mode\{1,2,3\} on Switchboard.}
\begin{tabular}{|c|c|c|}
\hline
criterion                   & K    & PPL  \\ \hline \hline
\multirow{2}{*}{NCE} & 100  & 54.4 \\ \cline{2-3} 
                       & 8000 & 52.7 \\ \hline
\multirow{2}{*}{Mode1} & 100  & 87.2 \\ \cline{2-3} 
                       & 8000 & 53.0 \\ \hline
\multirow{2}{*}{Mode2} & 100  & 53.9    \\ \cline{2-3} 
                       & 8000 &  53.5    \\ \hline
\multirow{2}{*}{Mode3} & 100  & 54.0 \\ \cline{2-3} 
                       & 8000 & 53.0 \\ \hline

\end{tabular}
\label{tab: swb_compare}
\end{table}

\begin{figure}[h!]
    \centering
    \includegraphics[scale=0.15]{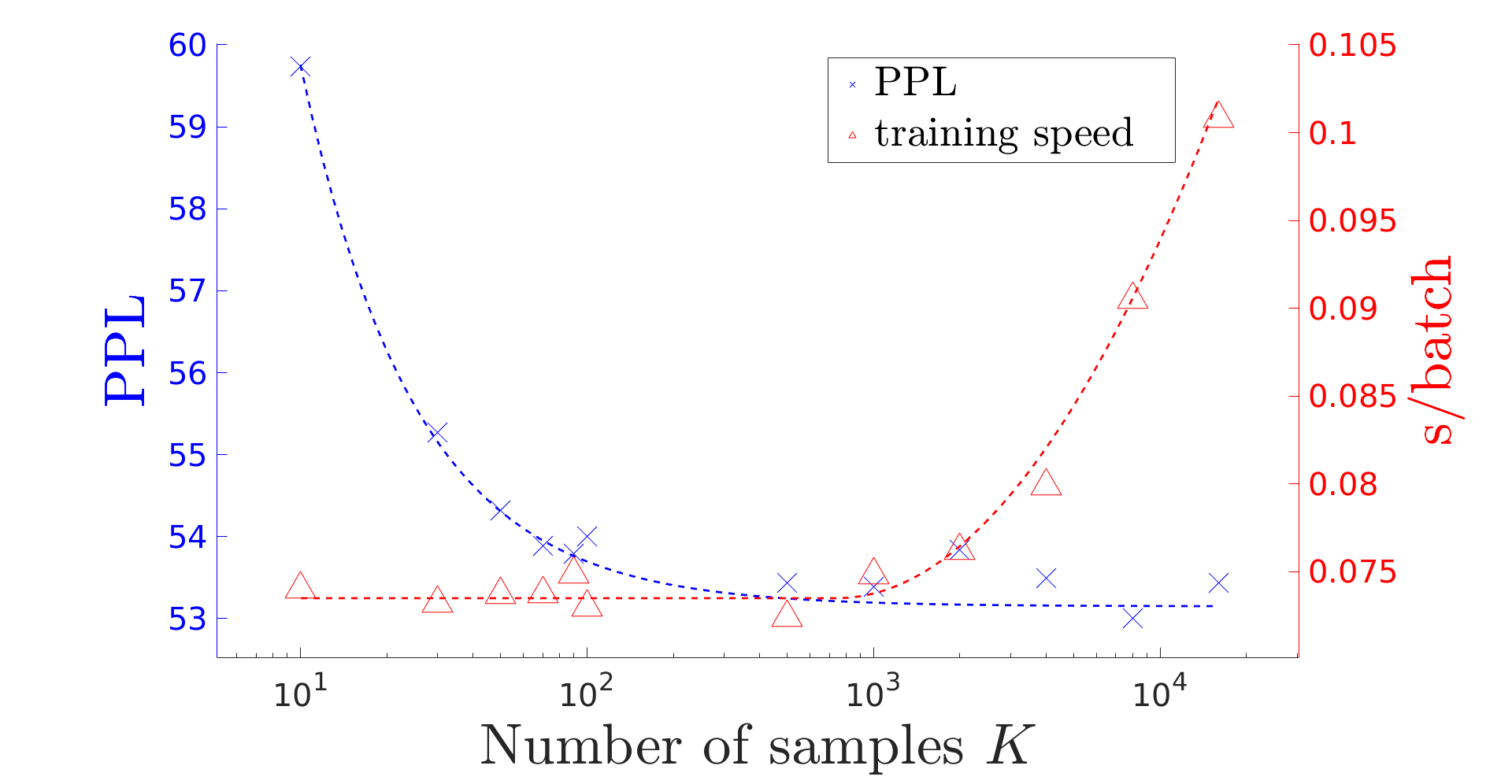}
    \caption{The influence of number of samples $K$.}
    \label{fig:swb_sweepk}
\end{figure}

We further investigate the influence of the number of samples $K$ on Switchboard in Mode3. Figure \ref{fig:swb_sweepk} shows that when $K$ is relatively small, the increase of $K$ almost does not influence the training speed, as the computation effort for the sum over $K$ is too small to influence the whole training procedure. When $K$ is large enough, increasing $K$ exponentially will slow down the training speed dramatically. On the other hand, when $K$ is relatively small, the increase of $K$ brings huge improvements in PPL. Because when $K$ is small, the sampling without replacement is similar to the sampling with replacement case, where the variance is proportional to $1/K$ and thus goes down fast with the increase of $K$, leading to a better approximation of the summation and fast decrease of PPL. When $K$ is large enough, the approximation is already good, so with the increase of $K$, the PPL only goes down slowly, and gradually converges. One can find the best trade-off between training speed and PPL according to the specific task. Surely, the trade-off also depends on the model architecture and the vocabulary size, and thus we do not do further experiments to investigate this relationship on other datasets.



Table \ref{tab: swbmainresult} shows the comparison of different criteria on Switchboard. In this task, LSTM LMs are interpolated with count-based LM for evaluation. In terms of PPLs, the CE baseline is slightly better than others, because the model outputs are well-normalized during training, which is more consistent with the evaluation on PPL. All other methods have similar performance. In terms of WER, CE baseline is slightly better, but the advantage is not large enough to draw the conclusion that CE outperforms sampling-based methods. In terms of training speed, sampling-based methods are faster than the traditional methods, since for sampling-based methods, the summation is over $K$, which is less than $C$. However, since the vocabulary size of Switchboard is not so large, the training speedup is limited. Compared to the original IS, the output of self-normalized IS (Mode3) can be used as scores directly in rescoring without correction, giving comparable training speed and performance.


\begin{table}[h!]
\center 
\caption{WERs of different criteria on Switchboard.}
\begin{tabular}{|c|c|c|c|c|c|}
\hline
\multirow{2}{*}{criterion} & \multirow{2}{*}{\begin{tabular}[c]{@{}c@{}}train time\\ (s/batch)\end{tabular}} & \multirow{2}{*}{PPL} & \multicolumn{3}{c|}{WER} \\ \cline{4-6} 
      &  &  & All & SWB & CH \\ \hline \hline
CE    & 0.100 & 49.9 &  10.1   &   6.8  &  13.4  \\ \hline
BCE   & 0.107 &52.3  &  10.3   &  6.9  &  13.7  \\ \hline
IS    & 0.079 & 51.5 & 10.3    & 7.0    &  13.7  \\ \hline
NCE   & 0.079 & 51.4 &  10.2   &  6.9   &  13.6  \\ \hline
Mode3 & 0.090 & 51.7 &   10.2  &   6.9  &  13.6  \\ \hline
\end{tabular}
\label{tab: swbmainresult}
\end{table}

Table \ref{tab: librimainresult} exhibits similar results on Librispeech. Since the vocabulary size is much larger than Swichboard, the training speedup gained from sampling is more significant, while the performance are still similar to traditional methods. With different model architectures and larger vocabulary sizes, we show that Mode3 is still competitive compared to NCE.

\begin{table}[h!]
\center
\caption{WERs of different criteria on Librispeech.}
\begin{tabular}{|c|c|c|c|c|}
\hline
\multirow{2}{*}{criterion} & \multirow{2}{*}{\begin{tabular}[c]{@{}c@{}}train time\\ (s/batch)\end{tabular}} & \multirow{2}{*}{PPL} & \multicolumn{2}{c|}{WER} \\ \cline{4-5} 
      &  &  & clean & other \\ \hline \hline
CE    &  0.302& 57.7 &  2.5     &    5.4   \\ \hline
BCE   & 0.358 & 58.5 &    2.5   &    5.4   \\ \hline
IS    & 0.206 & 58.4 &   2.6    &    5.5   \\ \hline
NCE   & 0.206 & 57.9 &   2.5    &    5.4   \\ \hline
Mode3 & 0.216 &58.3  &  2.5     & 5.4      \\ \hline
\end{tabular}

\label{tab: librimainresult}
\end{table}

Experimental results on EN are presented in Table \ref{tab: in_house_data}. Here, the baseline LM used for first-pass decoding is count-based. With LSTM LM rescoring, we observe a great improvement in WER, which is consistent with the common knowledge about neural network LMs outperforming count-based LMs for ASR.
 
Besides, we observe that Mode3 has similar performance to NCE for both $K=100$ and $K=8000$ on the production-oriented ASR dataset, validating our claim that the proposed self-normalized IS training criterion is a competitive sampling-based training criterion for neural LMs.

\begin{table}[]
\caption{WERs of different criteria on EN.}
\begin{tabular}{|c|c|c|c|c|c|}
\hline
corpus              & criterion              & K    & \begin{tabular}[c]{@{}c@{}}train speed\\ (s/batch)\end{tabular} & PPL & WER \\ \hline \hline
\multirow{5}{*}{EN} & baseline   & - &  - & 82.3 &13.7 \\ \cline{2-6} 
                    & \multirow{2}{*}{NCE}   & 100  &            0.092                                                &65.9&13.3  \\ \cline{3-6} 
                    &                        & 8000 &               0.098                                             &55.9&13.1 \\ \cline{2-6} 
                    & \multirow{2}{*}{Mode3} & 100  &               0.089                                             &68.0& 13.3 \\ \cline{3-6} 
                    &                        & 8000 &               0.114                                             &55.8&13.1\\ \hline 
\end{tabular}

\label{tab: in_house_data}
\end{table}

\section{Conclusion}

We propose self-normalized importance sampling for training neural language models.
Previously, noise contrastive estimation is a popular choice for sampling-based training criterion.
In our recent work, we show that other sampling-based training criteria, including importance sampling, can perform on par with noise contrastive estimation.
However, one caveat in that work is an additional correction step to recover the posterior distribution.
In this work, we completely negate the need for this step by modifying the importance sampling criterion to be self-normalized.
Through extensive experiments on both research-oriented and production-oriented datasets, we obtain competitive perplexities as well as word error rates with our improved method. 

\section{Acknowledgements}

This work has received funding from the European Research Council (ERC) under the European Union’s Horizon 2020 research and innovation programme (grant agreement No 694537, project ”SEQCLAS”). The work reflects only the authors’ views and none of the funding parties is responsible for any use that may be made of the information it contains. This work was partially supported by the project HYKIST funded by the German Federal Ministry of Health on the basis of a decision of the German Federal Parliament (Bundestag) under funding ID ZMVI1-2520DAT04A. We thank Christoph Lüscher for providing the LibriSpeech acoustic model and Eugen Beck for the lattices, and Markus Kitza for the SwitchBoard lattices.

\bibliographystyle{IEEEtran}

\bibliography{mybib}


\end{document}